\documentclass{article}

\usepackage[english]{babel}

\usepackage[letterpaper,top=2cm,bottom=2cm,left=3cm,right=3cm,marginparwidth=1.75cm]{geometry}

\usepackage{amsmath}
\usepackage{indentfirst}
\usepackage{graphicx}
\usepackage[colorlinks=true, allcolors=blue]{hyperref}

\title{Vision Transformer: Vit and its Derivatives}
\author{Zujun Fu}

\begin{document}
\maketitle

\begin{abstract}
Transformer, an attention-based encoder-decoder architecture, has not only revolutionized the field of natural language processing (NLP), but has also done some pioneering work in the field of computer vision (CV). Compared to convolutional neural networks (CNNs), the Vision Transformer (ViT) relies on excellent modeling capabilities to achieve very good performance on several benchmarks such as ImageNet, COCO, and ADE20k. ViT is inspired by the self-attention mechanism in natural language processing, where word embeddings are replaced with patch embeddings.

This paper reviews the derivatives in the field of ViT and the cross-applications of ViT with other fields.
\end{abstract}

\section{Pyramid Vision Transformer}

\begin{figure}
\centering
\includegraphics[width=0.8\textwidth]{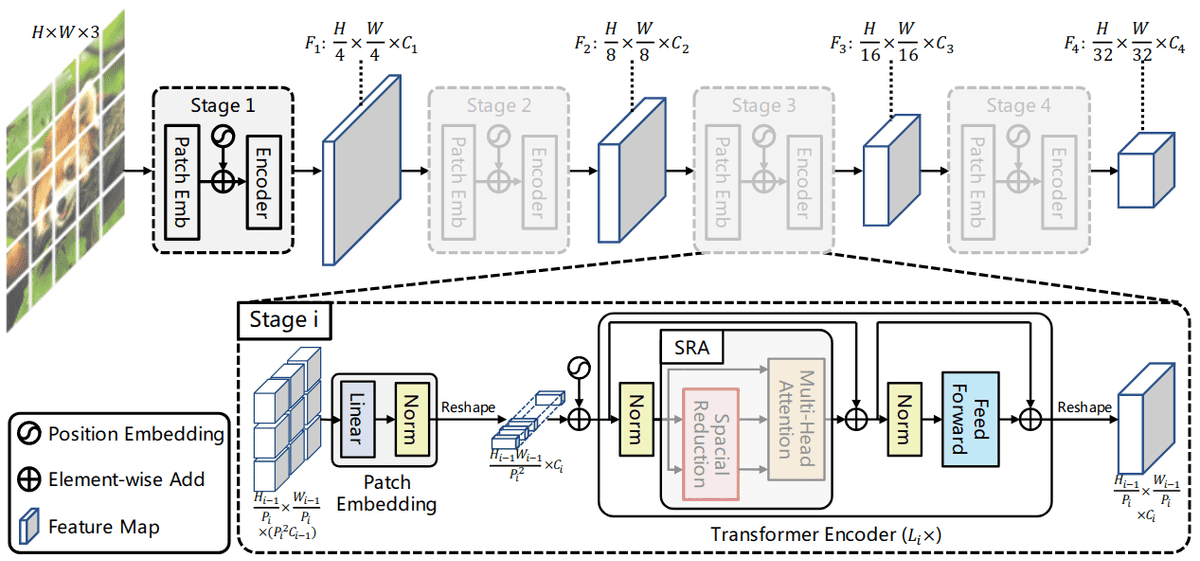}
\caption{Overall architecture of the proposed Pyramid Vision Transformer (PVT).}
\end{figure}

To overcome the quadratic complexity of the attention mechanism, the Pyramid Vision Transformer (PVT) uses a variant of self-attention called Spatial-Reduced Attention (SRA). It is characterized by spatial reduction of keys and values, similar to Linformer attention in the NLP field.

By applying SRA, the feature space dimension of the whole model is slowly reduced and the concept of order is enhanced by applying positional embedding in all transformer blocks.PVT has been used as a backbone network for object detection and semantic segmentation to process high resolution images.

Later on, the research team further improved their PVT model named PVT-v2, with the following major improvements.
\begin{itemize}
    \item overlapping patch embedding
    \item convolutional feedforward networks
    \item linear-complexity self-attention layers.
\end{itemize}

\begin{figure}
\centering
\includegraphics[width=0.9\textwidth]{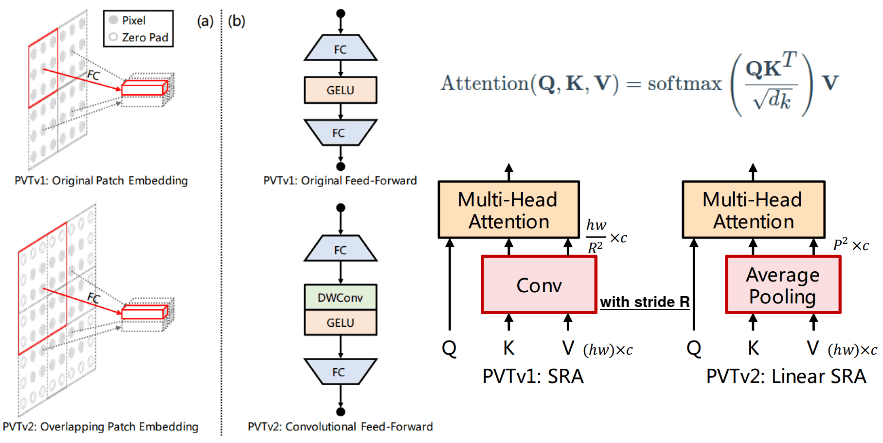}
\caption{PVT-v2.}
\end{figure}

Overlapping patches is a simple and general idea to improve ViT, especially for dense tasks (e.g. semantic segmentation).By exploiting overlapping regions/patch, PVT-v2 can obtain more local continuity of image representations.

Convolution between fully connected layers (FC) eliminates the need for fixed size positional encoding in each layer. The 3x3 deep convolution with zero padding (p=1) is designed to compensate for the removal of positional encoding from the model (they are still present, but only in the input). This process allows more flexibility to handle multiple image resolutions.

Finally, using key and value pooling(p=7), the self-attentive layer is reduced to a complexity similar to that of a CNN.

\section{Swin Transformer: Hierarchical Vision Transformer using Shifted Windows}

Swin Transformer aims to build the idea of locality from the standard NLP transformer, i.e. local or window attention:

\begin{figure}
\centering
\includegraphics[width=0.6\textwidth]{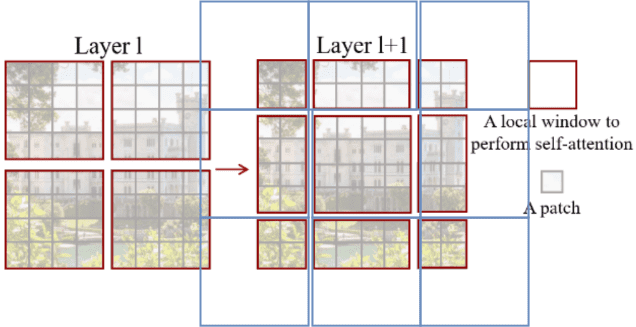}
\caption{Swin-transformer}
\end{figure}

In the Swin Transformer, local self-attention is used for non-overlapping windows. The next layer of window-to-window communication produces hierarchical representation by progressively merging windows.

\begin{figure}
\centering
\includegraphics[width=0.27\textwidth]{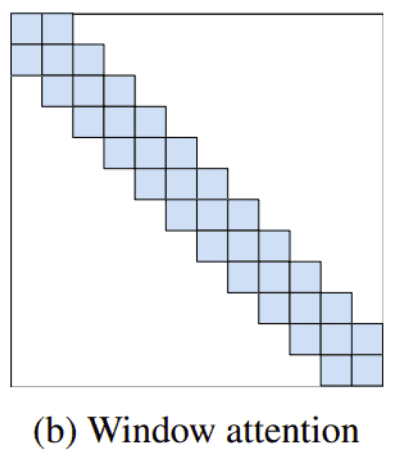}
\caption{local attention}
\end{figure}

As shown in \textbf{Figure 3}, the left shows the regular window partitioning scheme in the first layer, where self-attention is computed within each window. The window partitioning in the second layer on the right is shifted by 2 image patches, resulting in crossing the boundary of the previous window.

The local self-attention scales linearly with image size $O(M*N)$ instead of $O(N^2)$ in the window size used for sequence length $N$ and $M$.

By merging and adding many local layers, there is a global representation. In addition, the spatial dimensions of the feature maps has been significantly reduced. The authors claim to have achieved promising results on both ImageNet-1K and ImageNet-21K.

\section{Scaling Vision Transformer}

Deep learning and scale are related. In fact, scale is a key component in pushing the state-of-the-art. In this study, the authors from Google Brain Research trained a slightly modified ViT model with 2 billion parameters and achieved a top-1 accuracy of 90.45\% on ImageNet. This over-parameterized generalized model was tested on few-shot learning, with only 10 examples per class. A top-1 accuracy of 84.86\% was achieved on ImageNet.

Few-shot learning refers to fine-tuning a model with an extremely limited number of samples. The goal of few-shot learning is to motivate generalization by slightly adapting the acquired pre-trained knowledge to a specific task. If large models are successfully pre-trained, it makes sense to perform well with a very limited understanding of the downstream task (provided by only a few examples).

The following are some of the core contributions and main results of this paper.
\begin{itemize}
    \item Representation quality can be bottlenecked by model size, given that you have enough data to feed it;
    \item Large models benefit from additional supervised data, even over 1B images.
\end{itemize}

\begin{figure}
\centering
\includegraphics[width=0.45\textwidth]{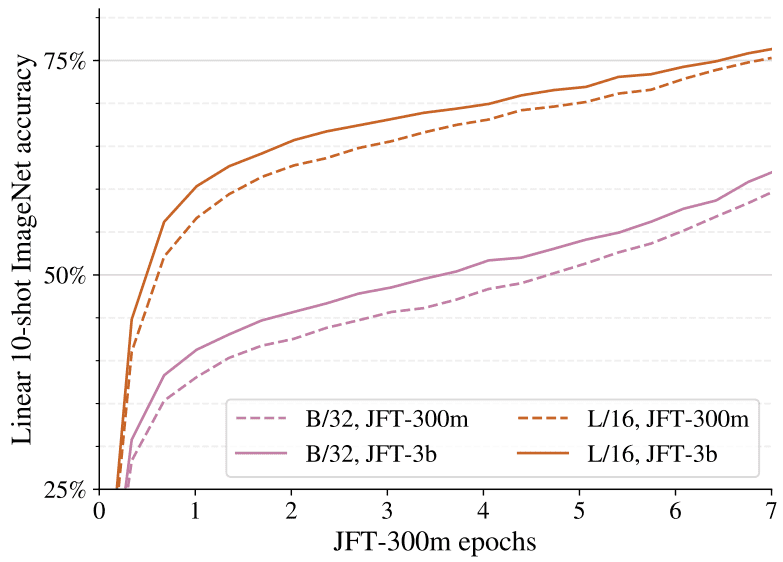}
\caption{scaling on jft data}
\end{figure}

\textbf{Figure 5} depicts the effect of switching from a 300M image dataset (JFT-300M) to 3 billion images (JFT-3B) without any further scaling. Both the medium (B/32) and large (L/16) models benefit from adding data, roughly by a constant factor. Results are obtained by few-shot(linear) evaluation throughout the training process.

\begin{itemize}
    \item Larger models are more sample efficient, achieving the same level of error rate with fewer visible images.
    \item To save memory, they remove class tokens (cls). Instead, they evaluated global average pooling and multi-head attention pooling to aggregate the representations of all patch tokens.
    \item They use different weight decay for the head and the rest of the layers called 'body'. The authors demonstrate this well in the \textbf{Figure 6}. The box values are few-shot accuracy, while the horizontal and vertical axes indicate the weight decay for the body and the head, respectively. Surprisingly, the stronger decay of the head produces the best results. The authors speculate that a strong weight decay of the head leads to representations with a larger margin between classes.
\end{itemize}

\begin{figure}
\centering
\includegraphics[width=0.6\textwidth]{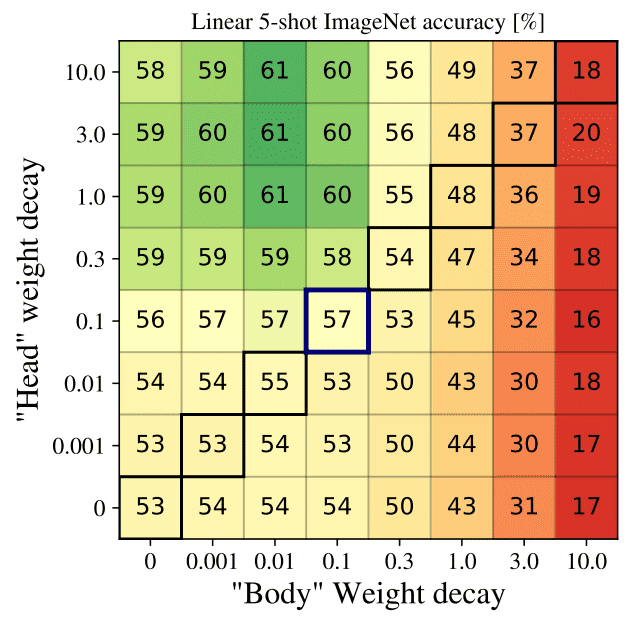}
\caption{Weight decay decoupling effect}
\end{figure}

This is perhaps the most interesting finding that can be more widely applied to pre-training ViT.

They used a warm-up phase at the beginning of training and a cool-down phase at the end of training, where the learning rate linearly anneals to zero. In addition, they used the Adafactor optimizer, which has a memory overhead of 50\% compared to traditional Adam.

\section{Replacing self-attention: independent token + channel mixing methods}

It is well known that self-attention can be used as an information routing mechanism with fast weights. So far, 3 papers tell the same story: replacing self-attention with 2 information mixing layers; one for mixing token (projected patch vector) and one for mixing channel/feature information.

\subsection{MLP-Mixer}

The MLP-Mixer contains two MLP layers: the first applied independently to the image patches (i.e., " mixing" the features at each location) and the other across the patches (i.e., " mixing" the spatial information).MLP Mixer architecture is shown in \textbf{Figure 7}.
\begin{figure}
\centering
\includegraphics[width=0.9\textwidth]{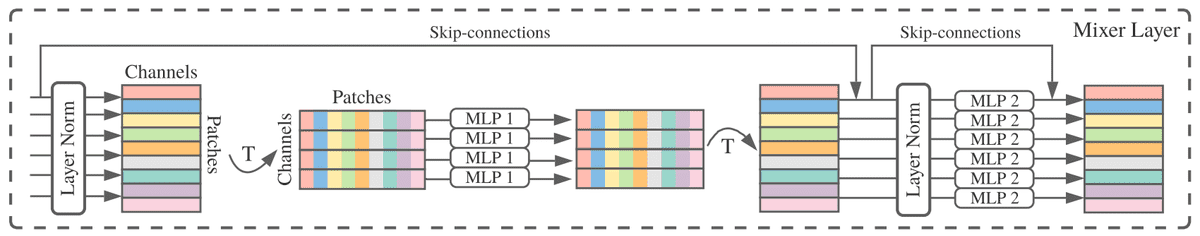}
\caption{MLP Mixer architecture}
\end{figure}

\subsection{XCiT: Cross-Covariance Image Transformers}

The other is the recent architecture XCiT, which aims to modify the core building block of ViT: self-attention applied to the token dimension.XCiT architecture is shown in \textbf{Figure 8}.
\begin{figure}
\centering
\includegraphics[width=0.9\textwidth]{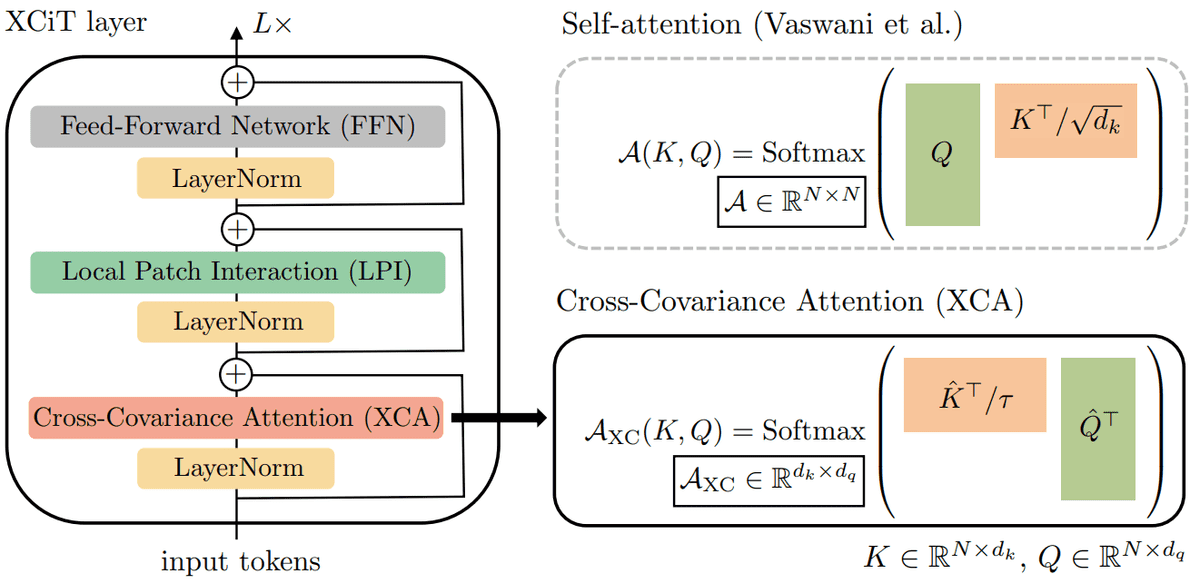}
\caption{XCiT architecture}
\end{figure}

\textbf{XCA}: For information mixing, the authors propose a cross-covariance attention (XCA) function that operates on the feature dimension of a token rather than on its own. Importantly, this method is only applicable to the L2-normalized set of queries, keys, and values. the L2 norm is denoted by the hat above the letters K and Q. The result of multiplication is also normalized to [-1,1] before softmax.

\textbf{Local Patch Interaction}: To achieve explicit communication between the patches, the researchers added two depth-wise 3×3 convolutional layers with Batch Normalization and GELU non-linearity in between, as shown in \textbf{Figure 9}. Depth-wise convolution was applied to each channel (here the patch) independently.

\begin{figure}
\centering
\includegraphics[width=0.9\textwidth]{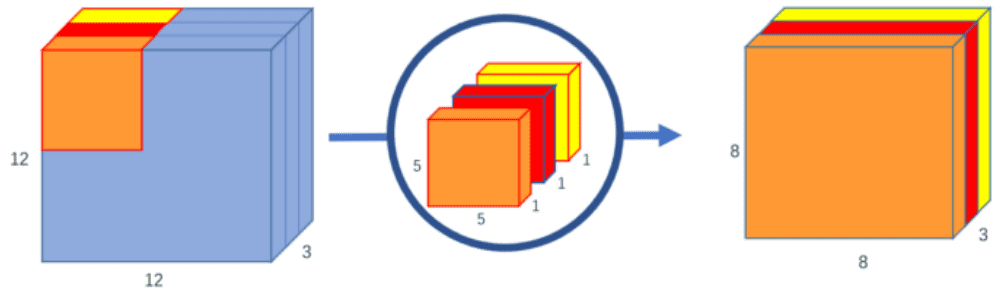}
\caption{depthwise convolutions}
\end{figure}

\subsection{ConvMixer}
Self-attention and MLP are theoretically more general modeling mechanisms, as they allow for larger receptive fields and content-aware behaviour. Nevertheless, the inductive bias of convolution has undeniable results in computer vision tasks.

Motivated by this, researchers have proposed another variant based on convolutional networks called ConvMixer, as shown in \textbf{Figure 10}. the main idea is that it operates directly on the patches as input, separating the mixing of spatial and channel dimensions and maintaining the same size and resolution throughout the network.

\begin{figure}
\centering
\includegraphics[width=0.9\textwidth]{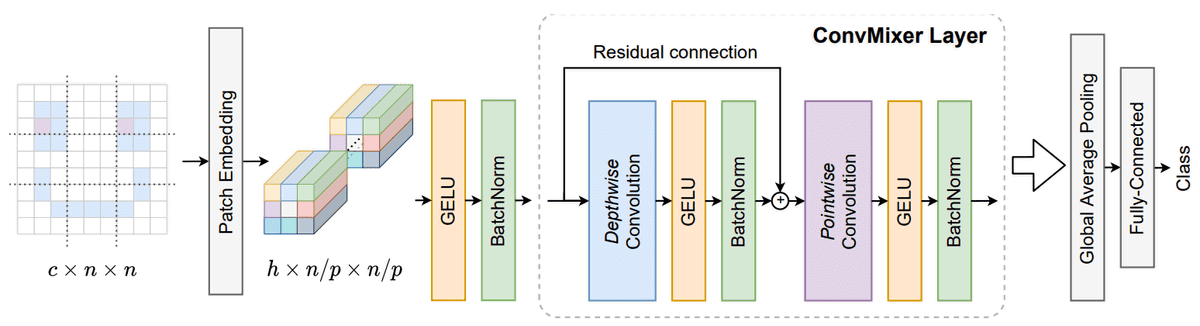}
\caption{ConvMixer architecture}
\end{figure}

More specifically, depthwise convolution is responsible for mixing spatial locations, while pointwise convolution (1x1x channel kernel) for mixing channel locations, as shown in the \textbf{Figure 11}.
\begin{figure}
\centering
\includegraphics[width=0.6\textwidth]{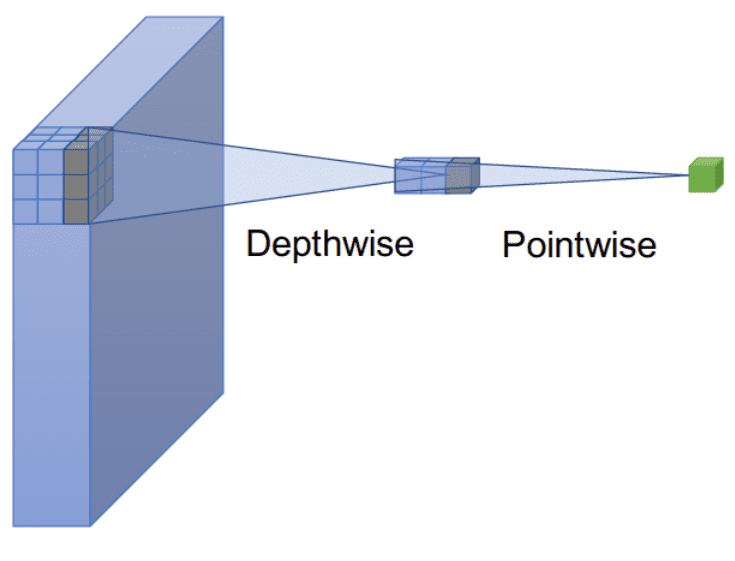}
\caption{depthwise convolution with pointwise convolution}
\end{figure}

Mixing of distant spatial locations can be achieved by selecting a larger kernel size to create a larger receptive field.

\section{Multiscale Vision Transformers}
The CNN backbone architecture benefits from the gradual increase of channels while reducing the spatial dimension of the feature map. Similarly, the Multiscale Vision Transformer (MViT) exploits the idea of combining a multi-scale feature hierarchies with a Vision Transformer model. In practice, the authors start with an initial image size of 3 channels and gradually expand (hierarchically) the channel capacity while reducing the spatial resolution.

Thus, a multi-scale feature pyramid is created, as shown in \textbf{Figure 12}. Intuitively, the early layers will learn high-spatial with simple low-level visual information, while the deeper layers are responsible for complex high-dimensional features. 
\begin{figure}
\centering
\includegraphics[width=0.88\textwidth]{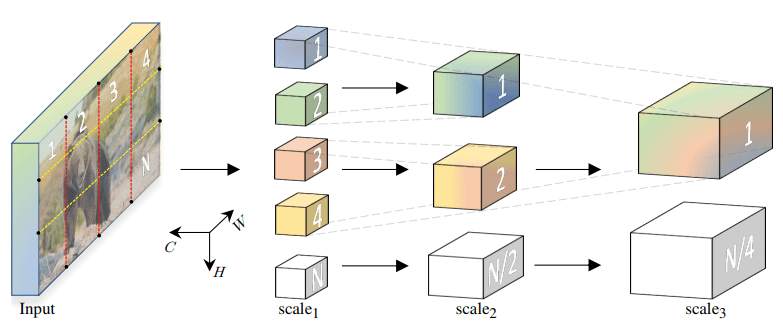}
\caption{Multi-scale Vit}
\end{figure}

\section{Video classification: Timesformer}
After a successful image task, the Vision Transformer is applied to video recognition. Two architectures are presented here,as shown in \textbf{Figure 13}.
\begin{figure}
\centering
\includegraphics[width=0.9\textwidth]{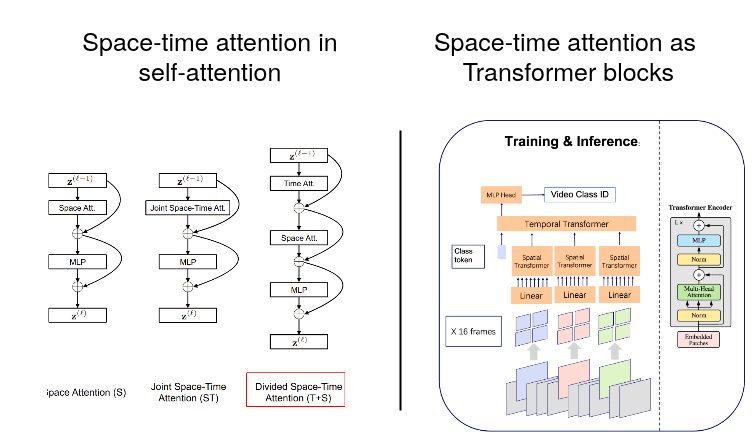}
\caption{Block-based and architecture-based / module-based space-time attention architectures for video recognition}
\end{figure}

\begin{itemize}
    \item Right: Reducing the architecture level. The proposed method applies a spatial Transformer to the projection image patches and then has another network responsible for capturing time correlations. This is similar to the winning strategy of CNN+LSTM based on video processing.
    \item Left: Space-time attention that can be implemented at the self-attention level, with the best combination in the red box. Attention is applied sequentially in the time domain by first treating the image frames as tokens. Then, the combined space attention of the two spatial dimensions is applied before the MLP projection. \textbf{Figure 14} is the t-SNE visualization of the method.
\end{itemize}

\begin{figure}
\centering
\includegraphics[width=0.9\textwidth]{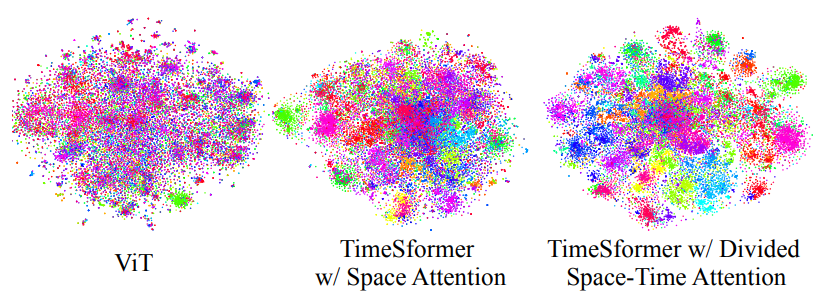}
\caption{Feature visualization with t-SNE of Timesformer}
\end{figure}

In \textbf{Figure 14}, each video is visualized as a point. Videos belonging to the same action category have the same color. A TimeSformer with split space-time attention learns more separable features semantically than a TimeSformer with only space attention or ViT.

\section{ViT in semantic segmentation: SegFormer}

NVIDIA has proposed a well-configured setup called SegFormer. SegFormer has an interesting design component. First, it consists of a hierarchical Transformer encoder that outputs multi-scale features. Second, it does not require positional encoding, which can deteriorate performance when the test resolution is different from the training. 

SegFormer ,as shown in \textbf{Figure 15}, uses a very simple MLP decoder to aggregate the multi-scale features of the encoder. Contrary to ViT, SegFormer uses small image patches, such as 4 x 4, which are known to favor intensive prediction tasks. The proposed Transformer encoder outputs 1/4, 1/8, 1/16, 1/32 multi-scale features at the original image resolution. These multi-level features are provided to the MLP decoder to predict the segmentation mask.
\begin{figure}
\centering
\includegraphics[width=0.9\textwidth]{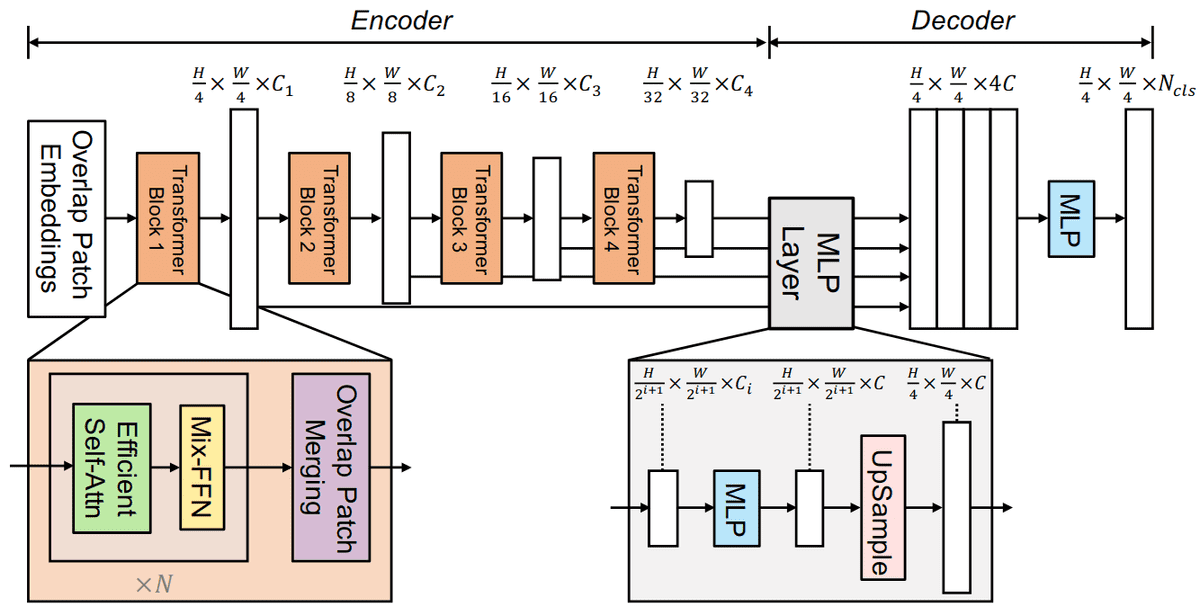}
\caption{Segformer architecture}
\end{figure}

Mix-FFN in \textbf{Figure 15}: In order to mitigate the impact of positional encoding, the researchers use zero-padding 3 × 3 convolutional layers to leak location information.Mix-FFN can be expressed as follows.
$$
x_{out}=MLP(GELU(Conv(MLP(x_{in}))))+x_{in}
$$

Efficient self-attention is proposed in PVT, which uses a reduction ratio to reduce the length of the sequence. The results can be measured qualitatively by visualizing the effective receptive field (ERF) as shown in \textbf{Figure 16}.
\begin{figure}
\centering
\includegraphics[width=0.87\textwidth]{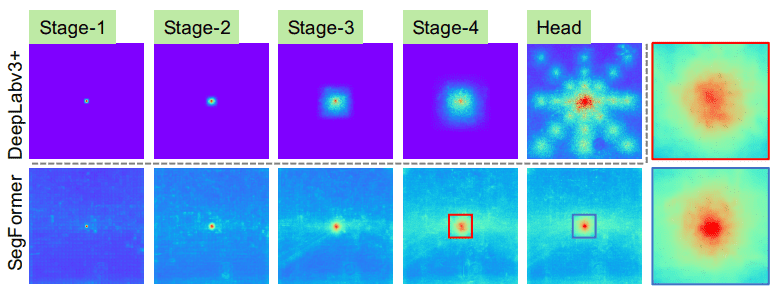}
\caption{SegFormer's encoder naturally produces local attention, similar to the convolution of lower stages, while being able to output highly non-local attention, effectively capturing the context of Stage-4. As shown in the enlarged patch, the ERF in the MLP header (blue box) differs from Stage-4 (red box) in that local attention is significantly stronger in addition to non-local attention.}
\end{figure}

\section{Vision Transformers in Medical imaging: Unet + ViT = UNETR}
Although there are other attempts in medical imaging, UNETR provides the most convincing results. In this approach, ViT is applied to 3D medical image segmentation. It was shown that a simple adaptation is sufficient to improve the baselines for several 3D segmentation tasks.

Essentially, UNETR uses the Transformer as an encoder to learn the sequence representation of the input audio, as in \textbf{Figure 17}. Similar to the Unet model, it aims to efficiently capture global multi-scale information that can be passed to the decoder through long skip connections, forming skip connections at different resolutions to compute the final semantic segmentation output.
\begin{figure}
\centering
\includegraphics[width=0.9\textwidth]{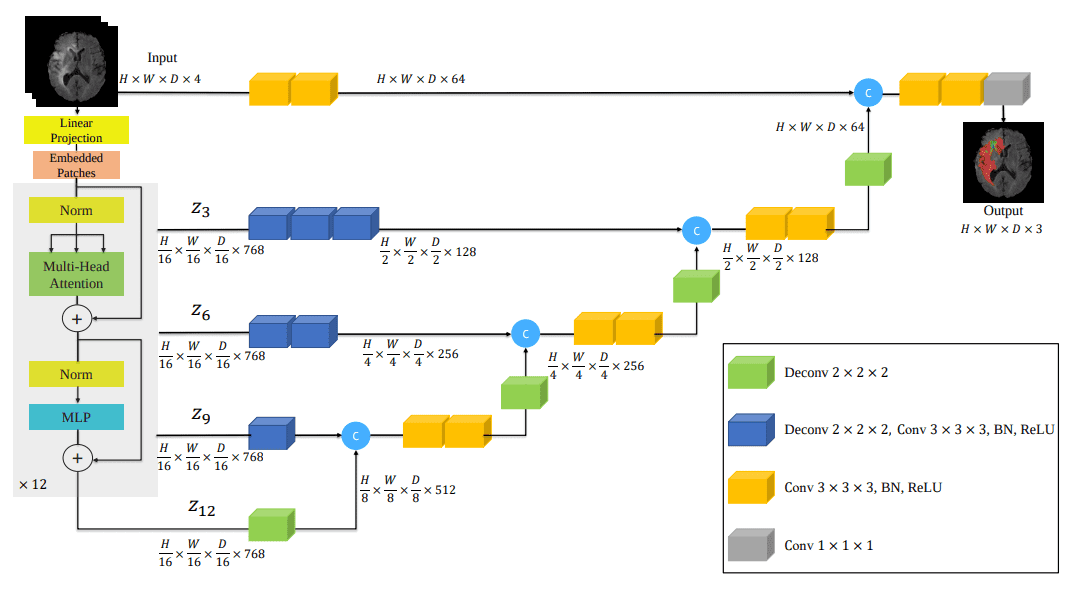}
\caption{Unetr architecture}
\end{figure}


\clearpage
\begin{thebibliography}{99}
	\bibitem{1} Wang, W., Xie, E., Li, X., Fan, D. P., Song, K., Liang, D., ... \& Shao, L. (2021). \textit{Pyramid vision transformer: A versatile backbone for dense prediction without convolutions}. arXiv preprint arXiv:2102.12122.
    \bibitem{2} Wang, S., Li, B. Z., Khabsa, M., Fang, H., \& Ma, H. (2020). \textit{Linformer: Self-attention with linear complexity}. arXiv preprint arXiv:2006.04768. 
    \bibitem{3} Wang, W., Xie, E., Li, X., Fan, D. P., Song, K., Liang, D., ... \& Shao, L. (2021). \textit{Pvtv2: Improved baselines with pyramid vision transformer}. arXiv preprint arXiv:2106.13797.
    \bibitem{4} Liu, Z., Lin, Y., Cao, Y., Hu, H., Wei, Y., Zhang, Z., ... \& Guo, B. (2021). \textit{Swin transformer: Hierarchical vision transformer using shifted windows}. arXiv preprint arXiv:2103.14030.
 	\bibitem{5} Zhai, X., Kolesnikov, A., Houlsby, N., \& Beyer, L. (2021). \textit{Scaling vision transformers}. arXiv preprint arXiv:2106.04560.
 	\bibitem{6} Tolstikhin, I., Houlsby, N., Kolesnikov, A., Beyer, L., Zhai, X., Unterthiner, T., ... \& Dosovitskiy, A. (2021). \textit{Mlp-mixer: An all-mlp architecture for vision}. arXiv preprint arXiv:2105.01601.
 	\bibitem{7} El-Nouby, A., Touvron, H., Caron, M., Bojanowski, P., Douze, M., Joulin, A., ... \& Jegou, H. (2021). \textit{XCiT: Cross-Covariance Image Transformers}. arXiv preprint arXiv:2106.09681.
 	\bibitem{8} \textit{Patches Are All You Need}? Anonymous ICLR 2021 submission
 	\bibitem{9} Fan, H., Xiong, B., Mangalam, K., Li, Y., Yan, Z., Malik, J., \& Feichtenhofer, C. (2021). \textit{Multiscale vision transformers}. arXiv preprint arXiv:2104.11227.
 	\bibitem{10} Bertasius, G., Wang, H., \& Torresani, L. (2021). \textit{Is Space-Time Attention All You Need for Video Understanding?}. arXiv preprint arXiv:2102.05095.
 	\bibitem{11} Xie, E., Wang, W., Yu, Z., Anandkumar, A., Alvarez, J. M., \& Luo, P. (2021). \textit{SegFormer: Simple and Efficient Design for Semantic Segmentation with Transformers}. arXiv preprint arXiv:2105.15203.
 	\bibitem{12} Hatamizadeh, A., Yang, D., Roth, H., \& Xu, D. (2021). \textit{Unetr: Transformers for 3d medical image segmentation}. arXiv preprint arXiv:2103.10504.
\end{thebibliography}
\end{document}